\documentclass{article}

    \PassOptionsToPackage{numbers, compress}{natbib}

\usepackage[final]{neurips_2022}

\usepackage[utf8]{inputenc} 
\usepackage[T1]{fontenc}    
\usepackage{hyperref}       
\usepackage{url}            
\usepackage{booktabs}       
\usepackage{amsfonts}       
\usepackage{nicefrac}       
\usepackage{microtype}      
\usepackage{natbib}
\usepackage[table,xcdraw]{xcolor}
\usepackage{wrapfig}
\usepackage{cleveref}
\crefformat{section}{\S#2#1#3} 
\crefformat{subsection}{\S#2#1#3}
\crefformat{subsubsection}{\S#2#1#3}

\newcommand{\mbar}{m-B-Ar}
\newcommand{\vbar}{B-Ar}
\newcommand{\tbar}{t-B-Ar}

\title{Parameter and Data Efficient Continual Pre-training for Robustness to Dialectal Variance in Arabic}

\author{%
Soumajyoti Sarkar \quad Kaixiang Lin \quad Sailik Sengupta \quad \\ \textbf{Leonard Lausen} \quad \textbf{Sheng Zha} \quad \textbf{Saab Mansour}\\
Amazon Web Services, USA \\
\texttt{\{soumajs,kaixianl,sailiks,lausen,zhasheng,saabm\}@amazon.com}\\
}

\begin{document}

\maketitle

\begin{abstract}
The use of multilingual language models for tasks in low and high-resource languages has been a success story in deep learning. In recent times, Arabic has been receiving widespread attention on account of its dialectal variance. While prior research studies have tried to adapt these multilingual models for dialectal variants of Arabic, it still remains a challenging problem owing to the lack of sufficient monolingual dialectal data and parallel translation data of such dialectal variants. It remains an open problem on whether the limited dialectical data can be used to improve the models trained in Arabic on its dialectal variants.  First, we show that multilingual-BERT (mBERT) incrementally pretrained on Arabic monolingual data takes less training time and yields comparable accuracy when compared to our custom monolingual Arabic model and beat existing models (by an avg metric of +$6.41$). We then explore two continual pre-training methods-- (1) using small amounts of dialectical data for continual finetuning and (2) parallel Arabic to English data and a Translation Language Modeling loss function. We show that both approaches help improve performance on dialectal classification tasks ($+4.64$ avg. gain) when used on monolingual models. 
\end{abstract}

\section{Introduction}
Pre-trained language models such as BERT \cite{devlin-etal-2019-bert} have been the backbone of many classification systems processing textual inputs. The two-step procedure for training these models is to first pre-train a language model $\mathcal{M}$ on some data followed by addition of a classification layer on top and finetuning $\mathcal{M}$ on a smaller target classification task data.
In multilingual scenarios, transformer-based architectures are trained on a large multilingual corpora at the pretraining stage \cite{devlin-etal-2019-bert,NEURIPS2019c04c19c2,pires2019multilingual}. Similar to mono-lingual models, they are then finetuned on a task specific to languages present in the pretraining data. The assumption that the finetuning task data has language overlap with the pretraining data is important; in zero-shot settings, continued pretraining on the new language is often necessary to improve model performance \cite{fujinuma2022match}.

In this paper, we seek to understand the distinctions between developing monolingual models from scratch and continual pretraining of multilingual models for Arabic, a language known to have rich dialectical variance.
While Modern Standard Arabic (MSA) has grained popularity at an international stage, dialectical variants like Egyptian Arabic, Levantine Arabic, and Moroccan Arabic are also widely used \cite{zaidan2014arabic}. Adapting to these dialectal variants is often non-trivial owing to differences between the dialects that may range from a syntactic (vocabulary, morphology) to a semantic one (same phrase could have different formality, context, or meaning) \cite{habash2013morphological}. This makes the task of developing Natural Language Systems for Arabic a challenging one. In light of these challenges, this paper discusses applying language model pretraining and incremental learning to tackle this resource deficiency of dialectal Arabic data. We expect our work to help motivate future work on developing models for dialectical Arabic and languages with dialectical diversity.


To complicate matters further, there is a distinct scarcity of large corpora for the dialectical variants in Arabic. It is not immediately evident if development of monolingual models vs. adaptation of multilingual models are the right path forward for scenarios where a language (in our case, Arabic) demonstrates dialectical diversity. For this purpose, we leverage and compare our models to publicly available monolingual araBERT \cite{antoun2020arabert} and multilingual BERT (mBERT) \cite{devlin-etal-2019-bert} models. 
To ensure our improvements don't come at the cost of impacting existing model performance, our benchmarks encompass both non-dialectical and dialectical tasks.


Our paper is structured as follows: we start off with describing the benchmarks, i.e. the downstream tasks that evaluate the different baseline and developed models on Arabic language and its dialectal variants (\cref{sec:benchmark}). We then focus on discussing methods to build our model, elucidating the model choices, the pretraining corpora and the objectives for adapting to dialectal data (\cref{sec:lmpretraining}). Finally, we demonstrate the performance of the various pretrained models on the downstream tasks.

\section{Downstream Task Benchmarks}
\label{sec:benchmark}

In order to evaluate the pretrained Arabic and multilingual models on dialectal variants, we use the eight tasks proposed in the ALUE benchmark \cite{seelawi2021alue}. This recent benchmark curates a diversified collection of Arabic NLU tasks that could be categorized as follows:  
\begin{enumerate}
    \item \textit{Single Sentence Classification}: It consists of 4 tasks on dialectal Arabic data - MADAR Shared Task Subtask 1 (MDD) which classifies sentences into their dialects based on cities, OSACT4 Shared Task on Offensive Language Detection (OOLD, where a tweet
    is labeled offensive if it consists of inappropriate
    language \& OHSD where offensive
    tweets from the same task are classified as hate speech based on a different criteria) and IDAT@FIRE2019 Irony Detection Task (FID).
    \item \textit{Sentence Pair Classification}: It consists of 2 tasks on NSURL-2019 Shared Task 8 (MQ2Q) in which a pair of questions is assumed to be classified to be similar if they have the same exact answer and meaning and  Cross-lingual Sentence Representations (XNLI) in which sentence pairs are labeled with either one of the following logical relationship labels: entailment, contradictory, or neutral.
    \item \textit{Multi-label classification}: SemEval-2018 Task 1 - Affect in Tweets, Emotion Classification task (SEC) on dialectal data.
    \item \textit{Regression}: Sentiment Intensity Regression task (SVREG) on dialectal data.
\end{enumerate}
Six out of the eight ALUE tasks contains dialectical data, of which five are constructed with twitter data, which by its very nature, has dialectical variance. This makes ALUE a suitable choice of evaluating our models since we experiment with adaptation to dialectal data.  More information on the ALUE data can be found in Appendix Section \ref{sec:alue_data_app}. We use the code script provided by the researchers \footnote{https://github.com/Alue-Benchmark/alue\_baselines}. Across the various tasks, we present Accuracy for XNLI, Pearson for SVREG and Jaccard for SEC, and F1-score for the other tasks. We use the same hyper-parameters used in the setup in \cite{seelawi2021alue}, and we finetune for 10 epochs with a maximum sequence length of 256 for all tasks.

\section{Language Model Pretraining} 
\label{sec:lmpretraining}

As mentioned earlier, we set out on training our own Arabic language model that exhibits robustness to dialectal variants. Owing to strict memory and inference-time constraints, we could not consider a BERT-Base (12-layer) model. Further, owing to limited task-time training data, we found a 4-layer transformer architecture to be the best-suited for our needs. For our comparison with baseline models, we truncate the publicly available BERT-Base monolingual AraBERT\footnote{https://github.com/aub-mind/arabert} and mBERT models by keeping only the first 4 layers of the 12 layer pretrained model. Although it has been known that different layers in BERT models capture different kinds of information \cite{jawahar2019does}, we also verified that truncating these publicly pretrained baseline models to 4-layers followed by task level fine-tuning did not degrade their performance significantly (as captured by the metrics in our benchmark compared to what is reported in the papers) while improving the training efficiency, inference latency, and memory requirements (as parameters reduce by one-third).

Exploring the setup of incremental learning with a 4 layer model instead of a 12 layer model allows us to measure the aspect of model adaptation to low resource dialectal data. We believe that any affirmation that smaller models can adapt to small amounts of dialectal Arabic even when pretrained on the dominant MSA data, could work as an indicator for a similar pattern for larger models on more data which we leave as future work. In addition to the publicly available AraBERT model, we explored training our own model from scratch due to (1) more independence in using pretrained dialectical data (eg. importance sampling, data-duplication strategies etc.) and (2) commercially restrictive licensing of AraBERT models.
The main questions we set out to answer in the course of training these models are:
\begin{itemize}
\item[RQ1] Will continually pretraining the multilingual models for fewer epochs bring comparable performance to a monolingual model trained from scratch when both are trained on the same corpora and the tokenizer vocab remains unchanged?
\item[RQ2] Given more dialectical data, does cost-effective continual finetuning of multilingual model perform at par with continual finetuning of monolingual models?
\item[RQ3] Can we leverage a parallel-data corpora to further boost model performance on?
\end{itemize}

\subsection{Pretraining corpora}
We use three different sets of pretraining corpora for the three stages of model training (in \cref{subsec:evolution}):

\begin{wraptable}{r}{0.52\textwidth}
\centering
\vspace{-5pt}
\begin{tabular}{lc}
\toprule
corpora                           & Size \\
\midrule
Oscar Arabic                     & 67M  \\
Arabic Wiki                      & 49M  \\
Arabic CC100                     & 111M \\
Arabic Newswire Part-1           & 2.3M \\
Arabic Gigaword Fifth Edition    & 96M  \\
Gulf Arabic Conv.                & 4K  \\ 
GALE (only Arabic data)          & 25K \\ 
BOLT SMS/Chat (only Arabic data) & 44K \\ 
\bottomrule
\end{tabular}
\caption{\footnotesize The number of sentences present in each corpora after a token-hashing heuristic was used to remove sentences with high duplicate tokens.}
\label{tab: pc_1}
\end{wraptable}

\textbf{[C1] Mixed Arabic and its dialectal variants\quad} The set composed of various Arabic corpora of different sizes as shown in ~\autoref{tab: pc_1}. Of these, the Arabic Wiki 2020 snapshot contains mixed English words as well as some transliterated sentences apart from the Arabic sentences. We use the Arabic source data of the GALE Arabic parallel data and the BOLT Egyptian Arabic SMS/Chat data, which is significantly smaller in size compared to the other corpora. We have $\approx$326 million sentences from these sources and use it to train models from scratch and continually pretrain m-BERT.

\textbf{[C2] Egyptian Arabic data\quad} We use the 2019 Egyptian Arabic dialectal corpora from Oscar; it contains 102K sentences. This data is used to continually pretrain a model already trained on \textit{C1} and gauge whether its inclusion helps in improving performance on ALUE. 

\textbf{[C3] Parallel Data\quad} We use parallel data ($\approx 255K$ sentences) on dialectical variants to English from the two GALE Arabic Parallel datasets, which contains transcripts for various news and conversations television programs broadcast in different dialects \cite{mzs07,mzs08}, and the BOLT Egyptian Arabic SMS/Chat dataset, which contains manual translations of messages in Egyptian Arabic \cite{t21}.

For deduplicating the sentences, we use a simple approximation method on each corpus separately in Table~\ref{tab: pc_1}. We space tokenize each sentence and for rolling windows of size 10 (in units of tokens) in a sentence, we compute the MD5 hash of each of these windows. Using these hash values, we compute the frequency of all such windows in all the sentences in a corpus. Following this, for each sentence, we compute the ratio of windows in a sentence which have been repeated more than one in the entire corpus. If that ratio is sufficiently high (we set that to 0.7), we remove that sentence from the corpus. To make sure that we do not remove all candidates among a pool of exactly duplicate sentences (the ratio would be 1 in that case for the exactly duplicate sentences), we keep one among the pool and remove the rest.

\subsection{Model evolution}
\label{subsec:evolution}

All models we trained are encoder models, specifically the BERT base architecture. As stated above, owing to resource and efficiency constraints, we use 4 layers for purposes of our experiments which amount to a total of 11M parameters. We use a global batch size of 1024 and a learning rate of 1e-5 in a multi-node cluster setup. Details on the hyperparamters for the pretraining is discussed in Appendix Section~\ref{sec:train_hparams}. The first three models we trained were the following:
\begin{itemize}
    \item[\mbar] Multilingual BERT (mBERT) continually pretrained for 800K steps starting from the pretrained public checkpoint
    \item[\vbar] 4-layer BERT model trained from scratch (with randomly initialized weights) with C1 for 1.3M steps
    \item[\tbar] \vbar model trained with a custom tokenizer trained on C1
\end{itemize}

 These three models were reasonably straightforward choices to compare with the publicly available AraBERT model\footnote{AraBERTv0.2-base: https://huggingface.co/aubmindlab/bert-base-arabert}. We trained \mbar for 800K steps and \vbar for 1.3M steps as we found the models started to overfit on the data as evidenced by the increase in validation losses after these many steps on the respective models. We further consider a version of AraBERT trained on Twitter Data (AraBERT-Twitter)\footnote{
AraBERTv0.2-twitter: https://huggingface.co/aubmindlab/bert-base-arabertv02-twitter} that we believe will exhibit dialectical robustness consider the dialectically diverse twitter corpora. For \mbar~and \vbar~, we use off-the-shelf multilingual bert tokenizer with the cased version of the vocab. For all the three models, we use the Masked Language Modeling (MLM) objective at the pretraining stage and discard the Next Sentence Prediction (NSP) unsupervised task. For \tbar, we trained a word-piece tokenizer with vocabulary size of 64K.  We present the results in \autoref{tab: results_1}. We clearly see that in this resource and latency constrained setting, our models outperform all of the existing models-- \tbar's avg. metric of $71.91$ is significantly higher compared to the best-performing model AraBERT with an avg. metric of $65.50$. 
We find that \mbar~and \vbar~perform similar to each other; $68.76~vs.~ 68.99$ on avg. across all tasks. With 128 NVIDIA A100 GPUs, \mbar~takes approximately 60\% of the time it takes to train \vbar~to convergence keeping all hyper-parameters the same. Hence, RQ1 holds.


\begin{table}[!t]
\begin{tabular}{l
>{\columncolor[HTML]{E7FAE6}}c
>{\columncolor[HTML]{E7FAE6}}c c
>{\columncolor[HTML]{E7FAE6}}c 
>{\columncolor[HTML]{E7FAE6}}c 
>{\columncolor[HTML]{E7FAE6}}c 
>{\columncolor[HTML]{E7FAE6}}c c}
\toprule
\textbf{Model}                     & \multicolumn{1}{l}{\textbf{FID}}   & \multicolumn{1}{l}{\textbf{MDD}} & \multicolumn{1}{l}{\textbf{MQ2Q}} & \multicolumn{1}{l}{\textbf{SVREG}} & \multicolumn{1}{l}{\textbf{SEC}}   & \multicolumn{1}{l}{\textbf{OOLD}} & \multicolumn{1}{l}{\textbf{OHSD}} & \multicolumn{1}{l}{\textbf{XNLI}} \\ \midrule
AraBERT         & 78.31                                                       & 51.15                                                     & 77.41                              & 42.41                                                       & 32.21                                                       & 94.92                                                      & 96.57                                                      & 51.02                              \\
AraBERT-Twitter & \cellcolor[HTML]{E7FAE6}79.73                               & \cellcolor[HTML]{E7FAE6}52.26                             & 77.07                              & \cellcolor[HTML]{E7FAE6}39.25                               & \cellcolor[HTML]{E7FAE6}31.34                               & \cellcolor[HTML]{E7FAE6}94.21                              & \cellcolor[HTML]{E7FAE6}97.76                              & 39.71                              \\
mBERT           & 77.14                                                       & 49.31                                                     & 77.11                              & 34.70                                                       & 35.49                                                       & 94.13                                                      & 96.49                                                      & \textbf{51.08}                     \\
\cmidrule{2-9}
\mbar                      & \cellcolor[HTML]{E7FAE6}79.61                               & \cellcolor[HTML]{E7FAE6}\textbf{56.04}                    & 80.26                              & \cellcolor[HTML]{E7FAE6}50.82                               & \cellcolor[HTML]{E7FAE6}41.05                               & \cellcolor[HTML]{E7FAE6}94.62                              & \cellcolor[HTML]{E7FAE6}97.13                              & 50.57                              \\
\vbar                       & \cellcolor[HTML]{E7FAE6}79.32                               & \cellcolor[HTML]{E7FAE6}55.84                             & \textbf{80.35}                     & \cellcolor[HTML]{E7FAE6}51.65                               & \cellcolor[HTML]{E7FAE6}41.88                               & \cellcolor[HTML]{E7FAE6}94.58                              & \cellcolor[HTML]{E7FAE6}97.27                              & 51.04                              \\
\tbar                      & \multicolumn{1}{l}{\cellcolor[HTML]{E7FAE6}\textbf{81.04}} & \multicolumn{1}{l}{\cellcolor[HTML]{E7FAE6}53.49}        & 72.63                              & \textbf{74.37}                                              & \multicolumn{1}{l}{\cellcolor[HTML]{E7FAE6}\textbf{49.26}} & \textbf{95.12}                                             & \textbf{98.36}                                             & 51.03                              \\
\bottomrule
\end{tabular}

\caption{\footnotesize Our models (last 3 rows) outperform existing models on all ALUE task metrics (Pearson for SVREG, Jaccard for SEC, accuracy for XNLI, and F1-score for the rest) except one. The columns in green denote benchmarks with dialectical data. All these models have 4 layers in the BERT style architecture.}
\label{tab: results_1}
\end{table}

\subsection{Training with Dialectal and Parallel Data}

While corpora C2 and C3 can be used along with C1, we observed higher gains in metrics when the dialectical data was used in the following ways:
\begin{itemize}
    \item[+C2] Continual pre-training with the dialectical corpora C2 and MLM loss.
    \item[+C2+C3] Continual pre-training of the above model using parallel data in C3 and a Translation Language Modeling (TLM) loss \cite{NEURIPS2019c04c19c2} where the two parallel sentences are concatenated and the position embedding is reset for the target sentence. 
\end{itemize}%
Owing to Out-Of-Vocabulary issues on a smaller fixed vocabulary, we ignore \tbar~for experiments in \autoref{tab:dialectical_results}. To answer RQ2, we show results of incrementally fine-tuning \mbar~and \vbar on dialectical data in \autoref{tab:dialectical_results}. First, we note that the performance of the \mbar~model hardly shows any improvement when dialectical data is used ($68.76 \rightarrow 69.16$ avg) compared to a noteworthy improvement seen on \vbar~($68.99 \rightarrow 73.14$ avg). \vbar~outperforms \mbar~models on six tasks and is within one F1-score/accuracy on the other two. Hence, RQ2 does not hold; continual finetuning of monolingual models trained on more dialectical data outperforms finetuning of multilingual models.

To answer RQ3, we noticed that the improvements are negligible when C3 and TLM loss is used for the multilingual \mbar~($69.16 \rightarrow 69.17$) and small when used on top of the monolingual \vbar~($72.94 \rightarrow 73.14$). On the six dialectical tasks, the improvement disappears for \mbar~($70.25 \rightarrow 69.96$) and slightly improves for \vbar~($74.45 \rightarrow 74.73$). On the dialectical benchmarks, we see gains when using \vbar+C2+C3 on almost all tasks (esp. high gains seen on MDD) but FID where a steep degradation averages out most of the gains seen. While further investigation and approaches to use the parallel corpora C3 is needed, we could not strongly conclude that use of a parallel data corpora was effective in boosting model performance.

\begin{table}[!t]
\centering
\begin{tabular}{l
>{\columncolor[HTML]{E7FAE6}}c
>{\columncolor[HTML]{E7FAE6}}c c
>{\columncolor[HTML]{E7FAE6}}c
>{\columncolor[HTML]{E7FAE6}}c
>{\columncolor[HTML]{E7FAE6}}c
>{\columncolor[HTML]{E7FAE6}}c c}
\toprule
\textbf{Model}      & \multicolumn{1}{l}{\textbf{FID}} & \multicolumn{1}{l}{\textbf{MDD}} & \multicolumn{1}{l}{\textbf{MQ2Q}} & \multicolumn{1}{l}{\textbf{SVREG}} & \multicolumn{1}{l}{\textbf{SEC}} & \multicolumn{1}{l}{\textbf{OOLD}} & \multicolumn{1}{l}{\textbf{OHSD}} & \multicolumn{1}{l}{\textbf{XNLI}} \\
\midrule
\mbar                      & \cellcolor[HTML]{E7FAE6}79.61                               & \cellcolor[HTML]{E7FAE6}56.04                    & 80.26                              & \cellcolor[HTML]{E7FAE6}50.82                               & \cellcolor[HTML]{E7FAE6}41.05                               & \cellcolor[HTML]{E7FAE6}94.62                              & \cellcolor[HTML]{E7FAE6}97.13                              & 50.57                              \\
\quad+C2    & 79.35                                                     & 56.60                                                     & 80.46                              & 53.72                                                       & 40.13                                                     & 94.56                                                      & 97.16                                                      & 51.33                              \\ 
\quad+C2+C3 & 78.29                                                     & 56.82                                                     & 80.65                              & 51.42                                                       & 40.75                                                     & \textbf{95.18}                                             & 97.51                                                      & \textbf{52.69}                     \\
\vbar                       & \cellcolor[HTML]{E7FAE6}79.32                               & \cellcolor[HTML]{E7FAE6}55.84                             & 80.35                     & \cellcolor[HTML]{E7FAE6}51.65                               & \cellcolor[HTML]{E7FAE6}41.88                               & \cellcolor[HTML]{E7FAE6}94.58                              & \cellcolor[HTML]{E7FAE6}97.27                              & 51.04                              \\
\quad+C2     & \textbf{81.20}                                            & 55.84                                                     & 84.73                              & 69.72                                                       & 47.66                                                     & 94.53                                                      & 97.75                                                      & 52.13                              \\ 
\quad+C2+C3  & 79.9                                                      & \textbf{57.61}                                            & \textbf{85.31}                     & \textbf{70.31}                                              & \textbf{48.03}                                            & 94.67                                                      & \textbf{97.91}                                                      & 51.38                              \\
\bottomrule
\end{tabular}
\vspace{5pt}
\caption{\footnotesize Effectiveness of dialectical and parallel data for continual pretraining.}
\label{tab:dialectical_results}
\end{table}


\vspace{-7pt}

\section{Conclusion}
\vspace{-7pt}
In this paper, we are able to show that existing multilingual models can be drastically improved with continual finetuning outperforming publicly available monolingual models on the ALUE benchmark. We observe that low-resource dialectical data can further improve performance of monolingual models, but don't have a pronounced effect on multilingual models. Unfortunately, parallel corpora data and Translation Language Modeling loss that can leverage this data does not yield further improvements. For future work, we also plan to investigate advanced alignment techniques and better leverage the parallel corpora data in the future. Additionally, while there is a broad range of studies in the area of adversarial training that could be applied to dialectal data, another area is to craft adversarial training examples using augmentations prior to training that can help adapt easily to dialectal data. In this paper, we considered Arabert as the baseline model for our comparison as it has variants trained on MSA as well as dialectal data and has been considered a robust baseline model until recently and in future, we plan to continue this work with comparisons on more recent SOA models on Arabic data which have been publicly released.


\medskip

\small

\bibliography{main}

\small

\appendix

\section{Appendix}

\subsection{Pretraining details} \label{sec:pretrain_params}

\subsubsection{Corpus}
The public links to the corpus used is below:

\begin{enumerate}
    \item OSCAR - \url{https://oscar-corpus.com/post/oscar-2019/}
    \item Arabic Wiki - \url{https://archive.org/details/arwiki-20190201}
    \item Arabic CC100 - \url{https://data.statmt.org/cc-100/}
    \item Arabic Gigaword Fifth Edition - \url{https://catalog.ldc.upenn.edu/LDC2011T11}
    \item Gulf Arabic Conversational Telephone Speech, Transcripts - \url{https://catalog.ldc.upenn.edu/LDC2006T15}
    \item GALE Phase 1 Arabic Broadcast News Parallel Text - Part 1 - \url{https://catalog.ldc.upenn.edu/LDC2012T18}
    \item GALE Phase 1 Arabic Blog Parallel Text - \url{https://catalog.ldc.upenn.edu/LDC2008T02}
    \item GALE Phase 1 Arabic Broadcast News Parallel Text - Part 2 - \url{https://catalog.ldc.upenn.edu/LDC2008T09}
    \item GALE Phase 1 Arabic Newsgroup Parallel Text - Part 1 - \url{https://catalog.ldc.upenn.edu/LDC2009T03}
    \item GALE Phase 1 Arabic Newsgroup Parallel Text - Part 2 - \url{https://catalog.ldc.upenn.edu/LDC2009T09}
    \item GALE Phase 2 Arabic Broadcast Conversation Parallel Text Part 1 - \url{https://catalog.ldc.upenn.edu/LDC2012T06}
    \item GALE Phase 2 Arabic Broadcast Conversation Parallel Text Part 2 - \url{https://catalog.ldc.upenn.edu/LDC2012T14}
    \item GALE Phase 2 Arabic Broadcast News Parallel Text - \url{https://catalog.ldc.upenn.edu/LDC2012T18}
    \item GALE Phase 2 Arabic Web Parallel Text - \url{https://catalog.ldc.upenn.edu/LDC2013T01}
    \item BOLT Egyptian Arabic SMS/Chat and Transliteration - \url{https://catalog.ldc.upenn.edu/LDC2017T07}
    \item BOLT Egyptian Arabic SMS/Chat Parallel Training Data - \url{https://catalog.ldc.upenn.edu/LDC2021T15}
Arabic Online Commentary Dataset - \url{https://aclanthology.org/P11-2007/}
\end{enumerate}

For processing the text, we take inspiration from some of the procedures adapted in AraBERT \footnote{https://github.com/aub-mind/arabert/} which includes removing emoji, tashkeel, tatweel,and html markup. Additionally, we also used some more steps to clean the Arabic wiki corpus. A lot of the sentences in the Arabic Wiki corpus contained templates from the page that specified the author name, the topic which did not contribute much to meaningful sentences. We removed such sentences using a combination of keyword search and regex expressions. We did not apply the Farasa pre-segmentation as done in \cite{antoun2020arabert} in any of corpus used to train our models. 

\subsection{Model Training} \label{sec:train_hparams}
We perform pre-training on 16 nodes, each with 8 NVIDIA A100 gpus. The distributed training was performed with Pytorch Lightning and DeepSpeed stage 2 \footnote{https://www.deepspeed.ai/tutorials/zero/} with full precision. We set the initial learning rate to 1e-5, with 10000 warmup steps. We used AdamW  \cite{loshchilov2017decoupled} optimizer with a learning rate linear decay, $\beta_1$ set to 0.9, $\beta_2$ set to 0.98, $\epsilon$ set to 1e-5, weight decay set to 0.2. We used a maximum norm of the gradients set to 0.1. We train keeping the maximum sequence length to 256.  As mentioned in the paper, we train 4-layer Multilingual BERT (mBERT) continually pretrained for 800K steps and 4-layer BERT model trained from scratch with C1 for 1.3M steps. We find that especially for 4-layer mBERT training beyond that with the corpus we had and the hyper-parameters leads to overfitting as observed from the validation loss. We use the same MLM masked loss strategy of 15\% masking but only setting 90\% of these 15\% to the MASK token and the rest 10\% to a random token.

\subsection{Finetuning details}
We use the open-source code for the ALUE benchmark \footnote{https://github.com/Alue-Benchmark/alue\_baselines/blob/master/bert-baselines/run\_alue.py} and all the default hyper-parameters set here. We set the maximum sequence length here to 256 and we train the model for 10 epochs.

\subsubsection{Evaluation on ALUE Data} \label{sec:alue_data_app}

\begin{enumerate}
\item IDAT@FIRE2019 Irony Detection Task (FID) : The shared task of Irony Detection in Arabic Tweets is based on a dataset of around 5,000 tweets. Each tweet is labeled with a "1" when it is ironic, holds satire, parody, sarcasm, or if the intended meaning is the contrary of the literal one. A label of "0" is given otherwise.

\item MADAR Shared Task Subtask 1 (MDD): Each sentence is exclusively classified into one of 25 labels, corresponding to one city out of 25 predefined Arab cities.

\item NSURL-2019 Shared Task 8 (MQ2Q):  In this task, a pair of questions is assumed to be semantically similar if they have the same exact answer and meaning, which is denoted with a label of "1". A label of "0" is given otherwise.

\item OSACT4 Shared Task on Offensive Language Detection (OOLD \& OHSD): For both of these tasks, the data contains 32,000 comments. We refine this corpus with 8 multi-labeled fine-grained classes, namely: toxic, insult, threat, identity hate, sexual, racial, blasphemy, and politically incorrect. Each label of these is denoted with either "1" if the class applies, or "0" otherwise.

\item SemEval-2018 Task 1 - Affect in Tweets (SVREG \& SEC): The first is the Emotion Classification task (SEC) in which a tweet is classified using one or more of eleven possible labels that best capture the emotions expressed by it. These labels are anger, anticipation, disgust, fear, joy, love, optimism, pessimism, sadness, surprise, and trust. The second is the Sentiment Intensity Regression task (SVREG) in which participants are expected to predict the "valence" of a given tweet, using a real-valued score between "0" and "1", with "0" representing the most negative sentiment possible, while "1" being the most positive sentiment possible.

\item Cross-lingual Sentence Representations (XNLI): These sentence pairs are labeled with either one of the following logical relationship labels: entailment, contradictory, or neutral. The data was originally labeled in the English language and then translated into 15 other languages including Arabic.

\end{enumerate}

\end{document}